\documentclass[onecolumn,runningheads]{llncs}
\usepackage{graphicx}
\usepackage{multicol}
\usepackage{multirow}
\usepackage{geometry}
\newgeometry{vmargin={18mm}, hmargin={22mm,17mm}}
\begin{document}
\title{"Forget" the Forget Gate: Estimating Anomalies in Videos using Self-contained Long Short-Term Memory Networks}
\titlerunning{"Forget" the Forget Gate}

\author{Habtamu Fanta\inst{}\orcidID{0000-0002-8394-8416} \and Zhiwen Shao\inst{}\orcidID{0000-0002-9383-8384} \and Lizhuang Ma\inst{}\orcidID{0000-0003-1656-4341}}
\authorrunning{H. Fanta et al.}

\institute{Shanghai Jiao Tong University, Shanghai, China  \\
	\email{\{habtamu\_fanta@,shaozhiwen@,ma-lz@cs.\}sjtu.edu.cn}}

\maketitle
\begin{abstract}
	Abnormal event detection is a challenging task that requires effectively handling intricate features of appearance and motion. In this paper, we present an approach of detecting anomalies in videos by learning a novel LSTM based self-contained network on normal dense optical flow. Due to their sigmoid implementations, standard LSTM's forget gate is susceptible to overlooking and dismissing relevant content in long sequence tasks like abnormality detection. The forget gate mitigates participation of previous hidden state for computation of cell state prioritizing current input. In addition, the hyperbolic tangent activation of standard LSTMs sacrifices performance when a network gets deeper. To tackle these two limitations, we introduce a bi-gated, light LSTM cell by discarding the forget gate and introducing sigmoid activation. Specifically, the LSTM architecture we come up with fully sustains content from previous hidden state thereby enabling the trained model to be robust and make context-independent decision during evaluation. Removing the forget gate results in a simplified and undemanding LSTM cell with improved performance effectiveness and computational efficiency. Empirical evaluations show that the proposed bi-gated LSTM based network outperforms various LSTM based models verifying its effectiveness for abnormality detection and generalization tasks on CUHK Avenue and UCSD datasets.
\end{abstract}
\section{Introduction}
Abnormal event detection is a hot research area that deals with identifying the presence of abnormal behaviour and possibly knowing its details from images or videos \cite{hinami2017joint}. Developing systems that can execute such abnormality identification tasks is important as the information gained can be used to assess the presence of any threat in an environment \cite{sun2017abnormal}. Modelling video data to extract meaningful anomalous features is challenging mainly because of the high dimension of videos and the presence of enormously interacting features across frames \cite{chong2017abnormal,kiran2018overview}. The widespread applicability of abnormal event detection in industry, academia and surveillance systems has attracted machine learning and computer vision researchers \cite{feng2016deep}.\par
Various abnormality detection methods rely on individually examining moving objects in a particular scene. For this, motion tracking and trajectory extraction techniques are usually employed to model peculiar activities and eventually single out anomalous events from the scene \cite{fu2005similarity,piciarelli2008trajectory}. In Wang et al. \cite{wang2006learning}, trajectories having spatial proximity manifesting related motion patterns are classified and used to identify outliers. Such tracking-based methods come short in abnormality detection performance when occlusion occurs amongst objects being tracked making them not well suited for studying intricate features in crowded environments \cite{xu2015learning,xu2017detecting}.\par
Though numerous attempts have been made to address abnormality detection problem from different perspectives, scarcity of large annotated dataset and the context-dependent nature of the problem domain are still decisive elements that need to be well addressed \cite{feng2016deep,ravanbakhsh2016plug,ravanbakhsh2017abnormal}. On top of these, effectively handling spatio-temporal features and complex long-term interactions between consecutive frames in videos is a further challenge \cite{ionescu2017unmasking,song2013one}. Even though deploying data augmentation techniques, developing generalizing models and employing various motion descriptors (features) have been proposed to tackle these challenges, there is yet a big gap to fill in the area.\par
Due to their inherent remarkable representation capability, deep convolutional neural networks have become a popular tool for modelling object segmentation and action recognition tasks. They have also shown to perform well for classification and prediction problems by learning on large set of data \cite{feng2016deep,hinami2017joint,sun2017abnormal}. Despite yielding tremendous results for such tasks, deep CNN's ability is constrained for learning spatio-temporal sequence data. Training deep CNNs on video data is also difficult because of the high dimensionality of videos. As deep CNNs are data-hungry (needing large set of data for training and testing), deploying them for problems with limited set of data (like abnormal event detection) is not a feasible solution. Abnormal event detection tasks deal with sequential video frames and the datasets are usually limited in size. Thus, deploying CNNs as core module for such task may not yield well learnt and representative models, which has recently paved way for the introduction of Recurrent Neural Networks (RNNs) \cite{chong2017abnormal}.\par
RNN in general and Long Short-Term Memory (LSTM) in particular have become popular frameworks for modelling long-term sequences. Over the years, the capability of LSTMs to contain the vanishing or exploding gradient issues of conventional RNNs has helped them to be adopted for sequence based tasks like language modelling, weather forecasting and speech recognition \cite{graves2013hybrid,sundermeyer2012lstm,xingjian2015convolutional}. They are also applied for abnormal event detection by coupling with convolution layers and autoencoders \cite{chong2017abnormal,feng2016deep,luo2017remembering}. Despite efforts to deploy LSTMs for abnormality detection, most methods focus on integrating them with autoencoders and variants of convolutional layers which limits exploiting the full potential of LSTMs. In addition, making LSTMs primary component of a network for modelling long-term sequences has not been thoroughly considered.\par
Even though the gating structures of LSTMs can contain vanishing or exploding gradients, they may squash important content in long-term sequences distorting sequential learning. In this paper, we address this problem by introducing a novel and light long short-term memory cell. We propose an end-to-end network built by stacking layers of the proposed LSTM cells capable of yielding enhanced performance on abnormal event detection benchmarks. During training, our network learns normal motion features from sparse and dense optical flow data independently. During testing, we evaluate the trained models on image-based video frames.\\
The major contributions of this work are summarized below:
\begin{itemize}
	\item We propose a novel, light long short-term memory architecture where we discard the forget gate and replace hyperbolic tangent activation with sigmoid function.
	\item We introduce a novel self-contained LSTM network built by stacking layers of the proposed LSTM.
	\item We show that the presented deep-LSTM network is computationally efficient and achieves effective detection performance over standard LSTM networks.
	\item Empirical evaluations also show that our light-LSTM based deep model performs better for generalization task over standard LSTMs on AED benchmarks.
\end{itemize}

The remainder of this paper is organized as follows. Section 2 reviews works related to abnormal event detection. Section 3 gives in-depth discussion on our methodology and the proposed network structure. Experimental analysis and results are discussed in Section 4, while Section 5 draws conclusions of the work.
\section{Related Work}
Abnormal event detection research has gained engrossing achievements over the past years where different approaches and techniques have been developed paving way for more investigation. A common approach of addressing abnormality detection tasks usually goes through two main phases. First, models are built by training on data containing only normal scenes. Then the normally trained model is tested and evaluated by supplying data containing mostly anomalous scenes \cite{hinami2017joint,ravanbakhsh2017abnormal,sun2017abnormal}. In this section, we give an overview of related works on abnormal event detection that focus on convolutional neural network and long short-term memory.

\subsection{CNN based methods}
As discussed in the previous section, deep convolutional neural networks have shown superior performance for object detection and action recognition tasks. They also yield better results than hand-crafted methods for abnormality detection task. An approach for abnormal event detection and recounting is introduced by Hinami et al. \cite{hinami2017joint}. In this work, generic knowledge is used to jointly detect abnormalities and identify their detailed attributes. This work proposes a model that can automatically identify anomalies from a scene without human intervention. Multi-task Fast R-CNN is utilized to detect visual concepts whose anomaly scores are then computed using one-class support vector machine, nearest neighbour and kernel density estimation detectors to measure the level of anomaly.\par
A spatio-temporal autoencoder that uses deep neural networks to automatically learn video representation is proposed by Zhao et al. \cite{zhao2017spatio}. Three-dimensional convolution layers are used by the deep network to extract spatial and temporal features in a better way. A novel weight-decreasing prediction loss is introduced to generate future frames and enhance motion feature learning.\par
An approach that decouples abnormality detection problem into a feature descriptor extraction appended by a cascade deep autoencoder (CDA) is introduced by Wang et al. \cite{wang2020abnormal}. The novel feature descriptor captures motion information from multi-frame optical flow orientations. Feature descriptors of the normal data are then provided as input for training the deep autoencoder based CDA network. During testing, the CDA model tries to reconstruct abnormal samples and accordingly estimate the reconstruction error.
\subsection{LSTM based methods}
Recent works have exploited LSTMs for abnormality detection task due to their ability to leverage sequential and temporal features. An unsupervised deep representation algorithm that couples stacked denoising autoencoders (SDAE) and LSTM networks is presented by Feng et al. \cite{feng2016deep}. Stacked denoising autoencoders are responsible for learning appearance and short-term motion cues while LSTMs keep track of long-term motion features to learn regularities across video frames. This work highlights the capability of LSTMs to better model long-term dependencies in video sequences.\par
A spatiotemporal network for video anomaly detection is presented by Chong et al. \cite{chong2017abnormal}. This work introduces a spatial encoder-decoder module populated with convolutional and deconvolutional layers, and a temporal encoder module made of convolutional LSTM layers. The spatial autoencoder handles spatial feature representation while the LSTM module learns sequential and temporal features.\par
Integrating convolutional neural network for appearance representation with convolutional LSTM for storing long-term motion information is introduced by Luo et al. \cite{luo2017remembering}. The developed ConvLSTM-AE architecture is capable of learning regular appearance and motion cues, and encoding variations in appearance and motion of normal scenes. The convolutional LSTM layer sits in between the convolutional and deconvolutional layers preserving spatial content of frames processed from previous layers and passes it to the next layer.\par
In addition to these CNN and LSTM based works, maintaining temporal coherency between video frames is shown to be effective for video processing tasks like human pose estimation \cite{liu2018human,xiao2014pose}. Liu et al. \cite{liu2018human} presented structured space learning and halfway temporal evaluation scheme for long-term consistency in videos. These works have shown the advantages of solving the temporal coherency problem in long video sequences. \par
Most of the endeavours put forth to apply LSTMs for abnormal event detection have coupled them with CNNs and autoencoders. In this paper, we investigate the potential of LSTMs (without fusing with other networks) for learning spatial and temporal features from appearance and motion scenes, and propose a novel, effective and efficient LSTM structure. The proposed LSTM is also capable of better maintaining temporal coherency of video frames.
\section{Overview of the proposed approach}
The LSTM-based model we present in this work (Figure 2) learns long-term, spatio-temporal sequential patterns and normal appearance and motion features from dense optical flow data. The network is built from six LSTM layers (each consisting of a varying number of our proposed LSTM unit). All LSTM layers except the last layer are appended by activation and batch normalization layers. We insert batch normalization layers in between these LSTM layers so as to obtain enhanced performance and computationally efficient network models \cite{ioffe2015batch}. While training deep networks, batch normalization tries to normalize the internal covariate shift resulting from uneven distribution of model features. It also enables trained models to have better generalization capability across different datasets.\par
\subsection{Input module}
The input module is built by grouping together \textit{T} (where \textit{T} is set to 4) consecutive dense optical flow frames in sliding window to generate a temporal cuboid. During preprocessing, we produce sparse and dense optical flows for every training video in CUHK Avenue and UCSD datasets. We convert these optical flow videos into frames and stack them together to form an input cuboid layer.\par
\subsection{Autoencoder module}
The autoencoder module emulates the function and structure of conventional autoencoders where we employ LSTM layers instead of convolutional and deconvolutional layers. We use 3 LSTM layers to build the encoding sub-module and 2 LSTM layers to make the decoding sub-module. Each LSTM layer is built based on the new LSTM architecture we propose in this work (discussed in Section 3.4) with varying number of units. The first and fifth LSTM layers are made from 32 LSTM units. The second and fourth LSTM layers consist of 16 LSTM units, while the third LSTM layer that acts as a bottleneck in the middle contains 8 LSTM units. The final LSTM layer is a single-unit layer that reduces the dimension of the previously learnt sequence back to a size compatible with the input cuboid. \par
During training, a model that is aware of the normal behaviour of a dataset (labelled \textit{'Model.h5'} in Figure 2) is generated. The model stores motion and appearance information about a normal environment. During testing, the trained model is supplied with test frames mostly consisting of anomalous scenes. The model tries to reconstruct the given frames where it fails on pixels containing anomalies.

\subsection{Sparse optical flow}
Sparse optical flow selects pixels that can be representative of an image or frame sequence. These representative pixels contain fairly enough content to present an image. Optical flow vectors keep track of these interesting pixels like corners and edges. Extracted features from one frame are sent to the next frame along a sequence to maintain consistency of pixels (features) under consideration. We adopt the Lucas-Kanade motion estimation technique to select and track the movement of interesting pixels in consecutive frames, which generates sparse optical flow vectors for our video data \cite{Lucas:1981:IIR:1623264.1623280}. This technique assumes that pixels in consecutive frames are not considerably far from each other and the time variable does not show noticeable increment between frames. It works by employing partial derivatives on spatial and temporal gradients to calculate the pixel flow at every location in an image \cite{nemade2019comparative}.

\subsection{Dense optical flow}
Dense optical flow features try to model and track motion information of every available pixel in a given image or frame sequence. Modelling motion cues with dense optical flow yields more precise result than sparse optical flow as the former considers all pixels in an image. Thus, it suits well for applications that require motion learning, video segmentation and semantic segmentation \cite{fischer2015flownet,ranjan2017optical}. In this work, we introduce a scheme of modelling normal behaviour by learning a network on dense optical flow vectors of a train set. We deploy the Gunnar Farneb{\"a}ck \cite{farneback2003two} method to generate dense optical flow vectors of videos in our train set. This method functions in a two-frame scenario by first employing quadratic polynomials to approximate the neighbourhoods of frames under consideration. It then applies a global displacement technique on these polynomials to build new signals. Finally, the global displacement is computed by using the coefficients yielded from the quadratic polynomials. The quadratic polynomial \textit{f(x)} is approximated in the local coordinate system using Eq.(\ref{eq:fx}).
\begin{equation}
\label{eq:fx}
f(x) \approx x\textsuperscript{T}Ax + b\textsuperscript{T}x + c
\end{equation}
where \textit{A} is a square matrix; \textit{b} and \textit{c} are vector and scalar variables respectively. These coefficients are approximated from the weighted least squares of signals in neighbourhood frames.\par
Based on the sparse and dense optical flow vectors generated using the Lucas-Kanade and Gunnar Farneb{\"a}ck methods respectively, we prepare train sets built from these optical flow vectors for each video in CUHK Avenue and UCSD train sets. By using a similar setup in the original datasets (i.e., equal number of frames per video) \cite{lu2013abnormal,mahadevan2010anomaly}, we convert these optical flow videos into same number of frames. Sample optical flows are presented in Figure 1.
\begin{figure}[!h]
	\centering
	\includegraphics[scale=1.0, width=120mm, height=90mm]{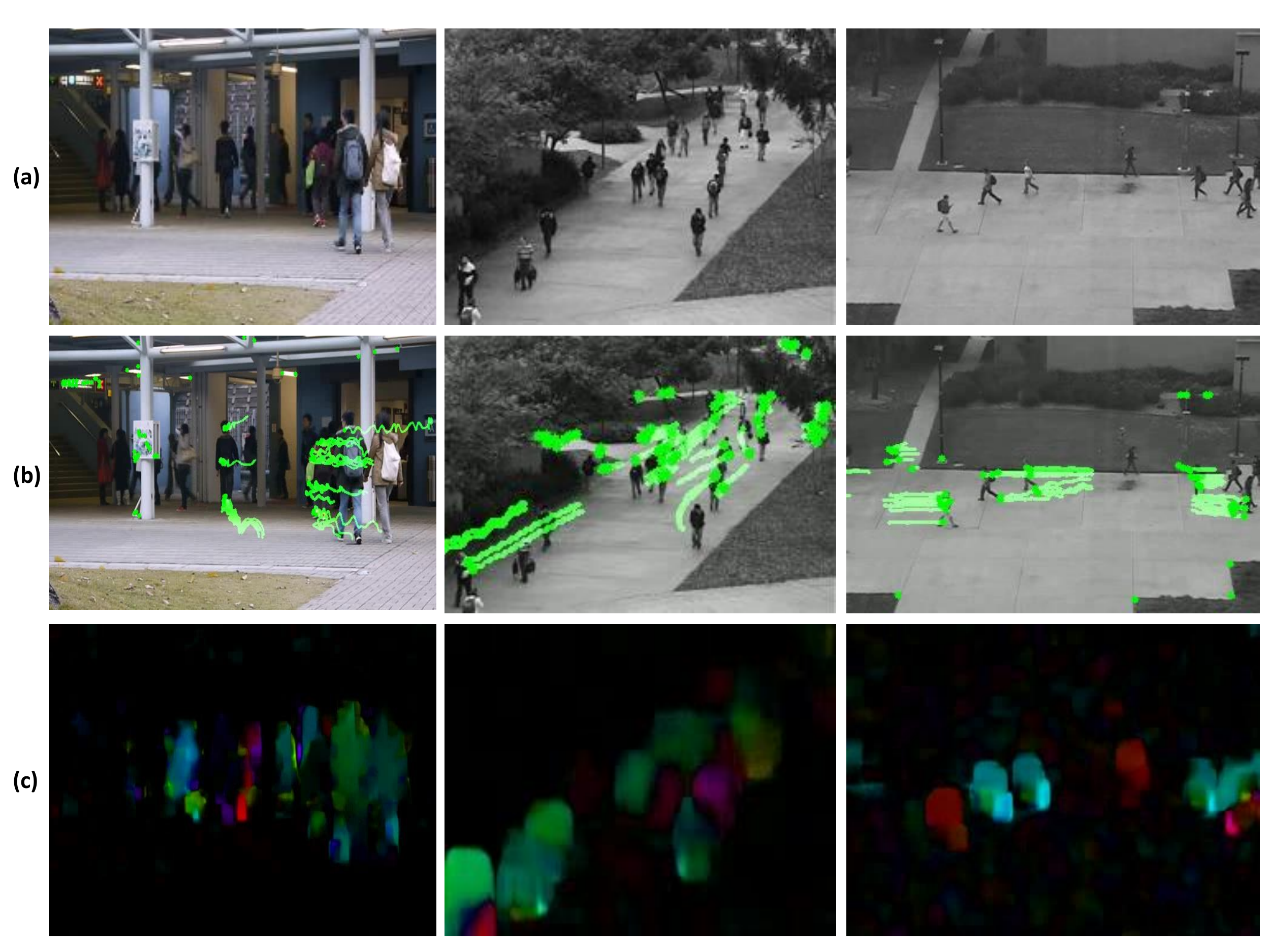}
	\vskip-0.2cm
	\caption{Sample frames from the original datasets, and the equivalent, preprocessed sparse and dense optical flows. \textbf{(a)} left-to-right: sample frames from Avenue, Ped1 and Ped2 train sets respectively. \textbf{(b)} left-to-right: sample sparse optical flows from Avenue, Ped1 and Ped2 train sets respectively. \textbf{(c)} left-to-right: sample dense optical flows from Avenue, Ped1 and Ped2 train sets respectively (appears best in colour).}
\end{figure}
\begin{figure*}[!h]
	\centering
	\includegraphics[scale=1.0, width=170mm, height=110mm]{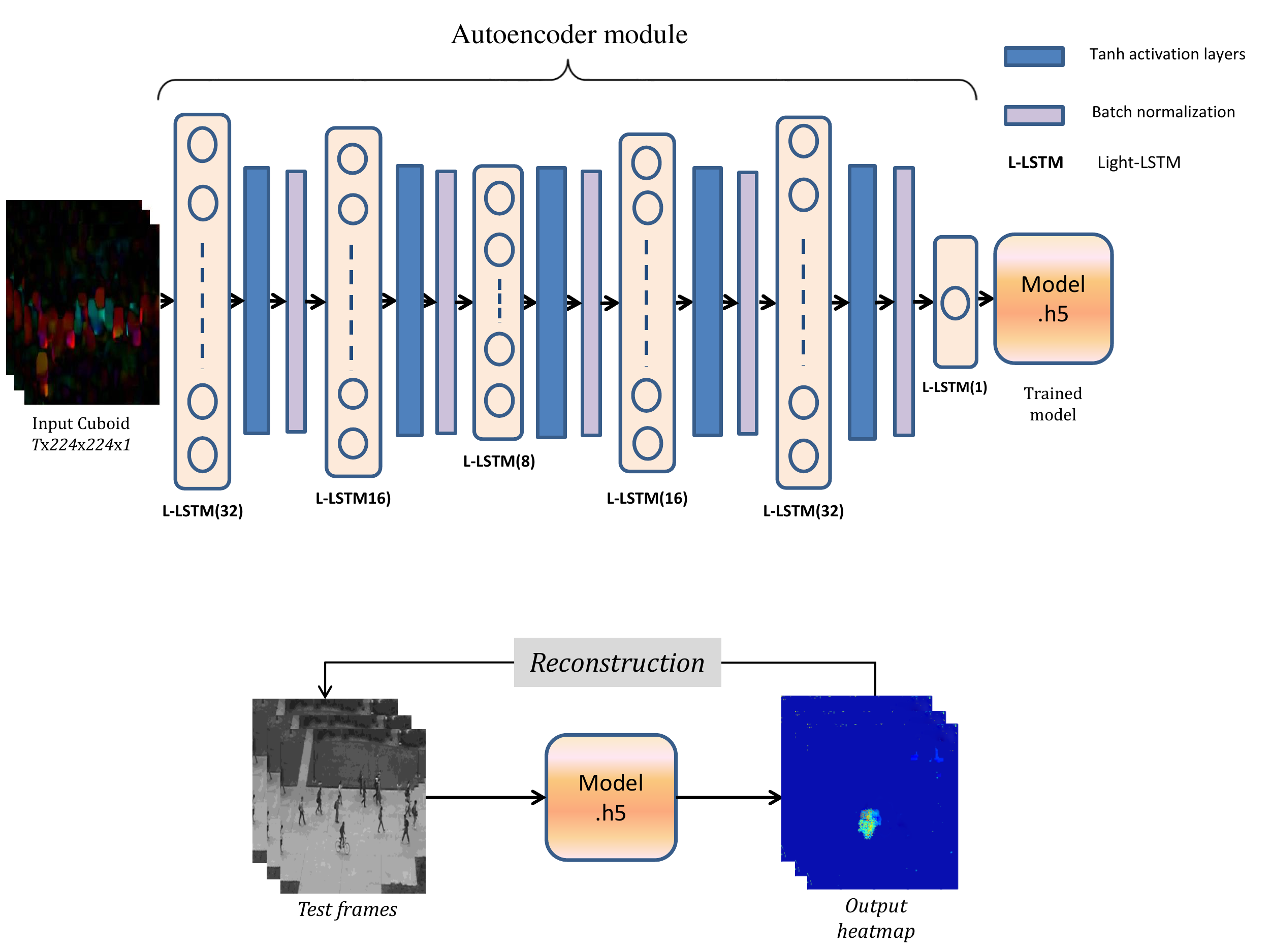}
	\vspace{-0.4cm}
	\caption{Architecture of the proposed self-contained sequential network and the subsequent testing process. \textbf{Top}: Training Phase - left to right: input layer (cuboid) made up of sequence of dense optical flow frames followed by an autoencoder module. The autoencoder module contains five LSTM layers (light-orange shaded) each appended by activation (blue-gray shaded) and batch normalization (light-rose shaded) layers, except the final layer. The vertical dashed lines in the L-LSTM layers indicate recurrent connections amongst the LSTM units \textbf{Bottom}: Testing Phase - left to right: sequence of video frames from test set fed to the trained model; a model file generated from the training phase; an output heatmap sequence produced by the trained model using input test frames (appears best in colour and zoom).}
\end{figure*}

\subsection{Long short-term memory}
Solving computer vision tasks like abnormality detection involves effectively modelling temporal interactions amongst inputs in long sequences. The ability of Recurrent Neural Networks (RNNs) to handle and manipulate such type of sequential data is proven to be better than conventional neural networks as they rely on state neurons to model context dependencies \cite{lyu2018road}. Structurally, RNNs contain a memory (state) with feedback or recurrent connections. A typical RNN is expressed in the following way:
\begin{equation}
	h_t=\delta(W\textsubscript{hx}x_t + W\textsubscript{hh}h\textsubscript{t-1} + b_h)
\end{equation}
\begin{equation}
	y_t=W\textsubscript{hy}h_t + b_y
\end{equation}
where \textit{h\textsubscript{t}} is the hidden state at time \textit{t}, \textit{$\delta$} is sigmoid activation function, \textit{W\textsubscript{hx}}, \textit{W\textsubscript{hh}} and \textit{W\textsubscript{hy}} are weight matrices, \textit{b\textsubscript{h}} and \textit{b\textsubscript{y}} are bias vectors, \textit{x\textsubscript{t}} is the input vector at time \textit{t}, and \textit{y\textsubscript{t}} is the output vector at time \textit{t}.
\begin{figure}[!h]
	\centering
	\includegraphics[scale=1.0, width=85mm, height=60mm]{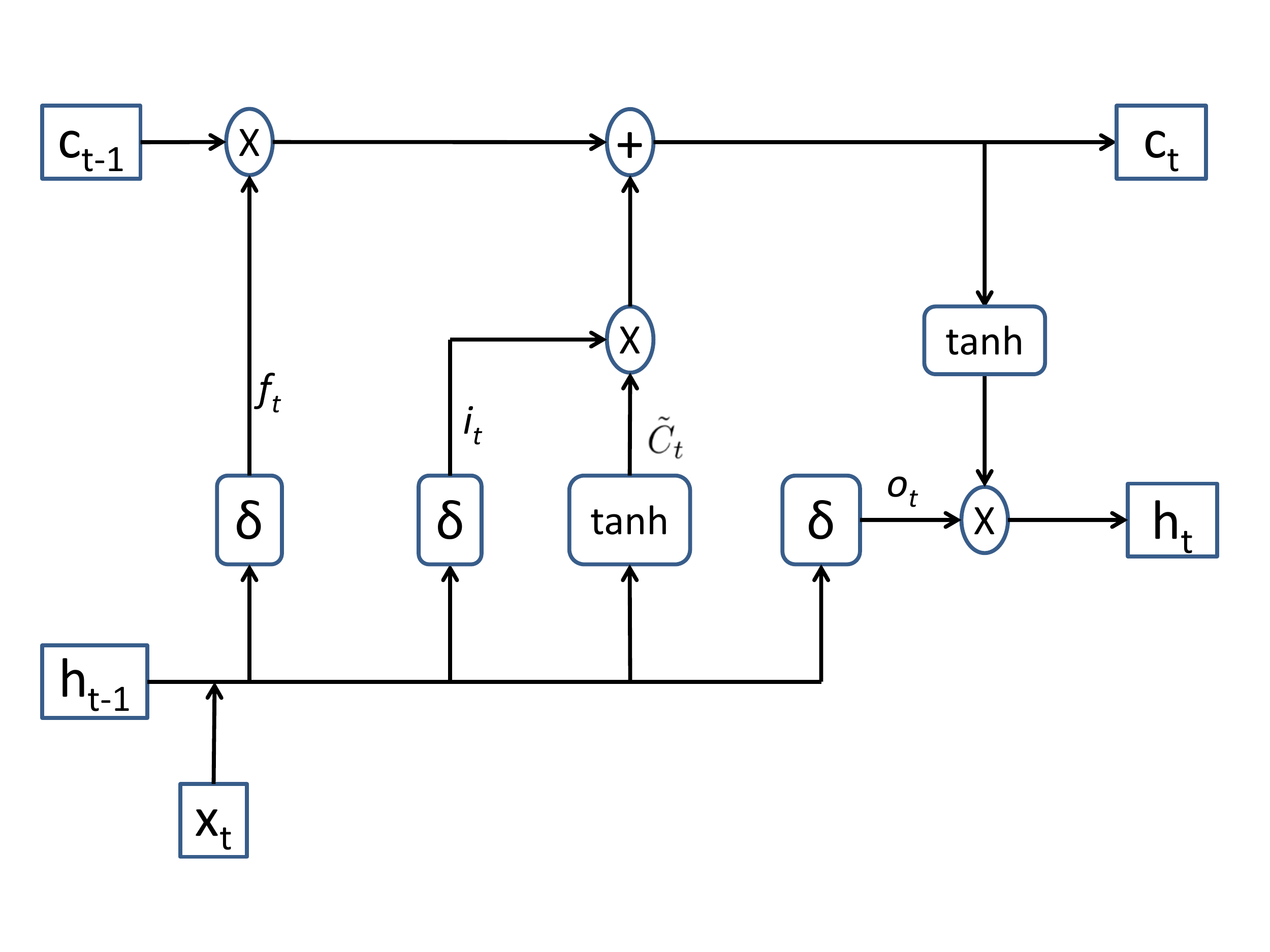}
	\vskip-0.7cm
	\caption{Structure of a standard LSTM cell.}
\end{figure}
Despite their effectiveness for learning long-term sequences, RNNs usually suffer from the vanishing or exploding gradient problem making their training process cumbersome. Long Short-Term Memory (LSTM), a variant of RNN, is capable of generating stable and robust models capable of representing long range interactions between events. LSTMs can successfully contain the vanishing or exploding gradient issue during back propagation with their inherent gating mechanisms \cite{chong2017abnormal,xingjian2015convolutional}.
An LSTM cell (structure depicted in Figure 3) possesses three gates (forget gate, input gate, output gate) that monitor the flow of information into and out of its memory. The forget gate, \textit{f\textsubscript{t}}, (also called recurrent gate) controls for how long a stored data from previous state should stay in memory. The input gate, \textit{i\textsubscript{t}}, is responsible for feeding new data to memory; the output gate, \textit{o\textsubscript{t}}, controls the impact of stored data on activation of the output block \cite{chong2017abnormal,sosa2017twitter}. The following formulae show the mathematical model of a typical LSTM cell:
\begin{equation}
	\label{eq:f_t}
	f_t=\delta(W_f\otimes[h \textsubscript{t-1},x_t] + b_f)
\end{equation}
\begin{equation}
	\label{eq:i_t}
	i_t=\delta(W_i\otimes[h \textsubscript{t-1},x_t] + b_i)
\end{equation}
\begin{equation}
	\label{eq:tilde_C_t}
	\tilde{C}_t =tanh(W_C\otimes[h \textsubscript{t-1},x_t] + b_C)
\end{equation}
\begin{equation}
	\label{eq:C_t}
	C_t=f_t\otimes C\textsubscript{t-1} + i_t\otimes \tilde{C}_t
\end{equation}
\begin{equation}
	\label{eq:o_t}
	o_t=\delta(W_o\otimes[h \textsubscript{t-1},x_t] + b_o)
\end{equation}
\begin{equation}
	\label{eq:h_t}
	h_t=o_t\otimes tanh(C_t)
\end{equation}

Eq.(4) handles the forget layer; new information is supplied to the model using Eq.(5) and Eq.(6); Eq.(7) unifies previous information with new one while Eq.(8) and Eq.(9) output the learning results and send it to the next time step LSTM unit. \textit{x\textsubscript{t}} is the input vector, \textit{h\textsubscript{t-1}} is the previous hidden state, \textit{h\textsubscript{t}} is the hidden state at time t, \textit{\~C\textsubscript{t}} is the candidate cell state, \textit{C\textsubscript{t}} is the cell state at time t, \textit{C\textsubscript{t-1}} is the previous cell state, \textit{W} denotes trainable weights in matrix form, \textit{b} are the bias vectors, and $\otimes$ denotes an element-wise multiplication (Hadamard product) operation.\par
Even though standard LSTMs are capable of entertaining long-term sequential data and solve the vanishing or exploding gradient problem of RNNs, their gating structures are prone to overlooking (missing) important content in a long sequence. Unless they are controlled, such gating structures may lead to ill-learnt models where invaluable contents and long-term dependencies are not well considered during training. We propose a mechanism that mitigates this risk of information loss by proposing a new LSTM architecture which is capable of equally treating important content in long sequences.

\subsection{Proposed LSTM architecture}
The modifications we employ on standard LSTM cells are fully removing the forget gate and substituting the hyperbolic tangent activation used for candidate cell state computation with logistic sigmoid activation. These alterations generate an effective and efficient, light-weight model. The new bi-gated LSTM architecture shows performance improvements on public abnormal event detection benchmarks over the conventional, tri-gated architecture.

\subsubsection{Removing the forget gate}
The forget gate in LSTMs (modelled in Eq.(\ref{eq:f_t})) decides how much of the information from previous hidden states should be removed or kept across a sequence. This enables LSTM-based models to learn which share of previous hidden states are relevant and should be carried to the next hidden state. This behaviour of forget gates is advantageous for scenarios where discontinuations are constantly observed in sequential data or for applications that positively overlook the significance of previous information by prioritizing current content. For these cases, the forget gate tries to nullify the impact of previous memory state while computing candidate memory by giving attention to current input.\par
For abnormal event detection problems that focus on analyzing appearance and motion patterns, an LSTM-based model should keep information from previous memory for longer duration to effectively compute candidate memory state. Sustaining such information from previous frames and learning their features enables models to reasonably compute the candidate state at every time step. This makes current memory state well informed of previous content and more reliable for performance evaluation. So, we propose removing the forget gate from LSTMs for abnormality detection as we need to keep all information from previous sequences. Empirically, removing the forget gate is accomplished by setting the variable \textit{f\textsubscript{t}} (in Eq.(\ref{eq:f_t})) to 1. In doing so, we are explicitly allowing the sigmoid function in the equation to attain a maximum value of 1, thereby not squashing any of the previous hidden states \textit{h\textsubscript{t-1}}.


\subsubsection{Replacing the hyperbolic tangent activation}
In addition to removing the forget gate, we also replace the hyperbolic tangent activation of candidate cell state (Eq.(\ref{eq:tilde_C_t})) with logistic sigmoid activation. Eq.(\ref{eq:modtildeC_t}) shows the candidate cell state computation with logistic sigmoid activation.
\begin{equation}
	\label{eq:modtildeC_t}
	\tilde{C}_t =\delta(W_C\otimes[h \textsubscript{t-1},x_t] + b_C)
\end{equation}
Hyperbolic tangent activation used in conventional LSTMs is not effective when learning feed-forward networks. The performance of such networks decreases significantly whenever the network goes deeper \cite{gulcehre2016noisy}.
\begin{figure}[!h]
	\centering
	\includegraphics[scale=1.0, width=80mm, height=60mm]{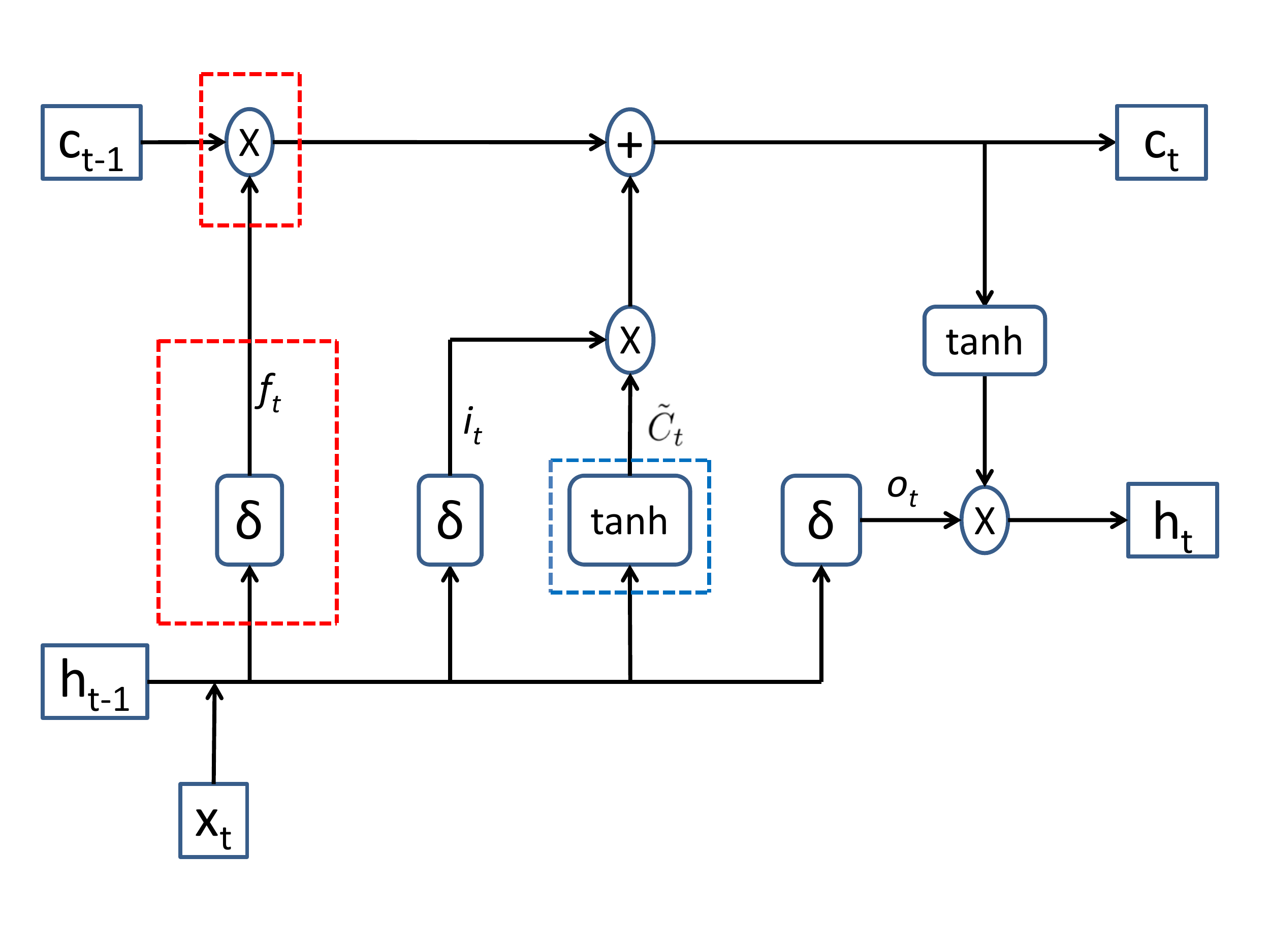}
	\vspace{-0.4cm}
	\caption{Alterations made to standard LSTM cell. The red-dashed rectangle shows the forget gate (with its associated point-wise multiplication operator) eliminated by our approach. The blue-dashed rectangle surrounds \textit{tanh(.)} activation replaced by \textit{sigmoid} activation (appears best in colour).}
\end{figure}

\begin{figure}[!h]
	\centering
	\includegraphics[scale=1.0, width=80mm, height=55mm]{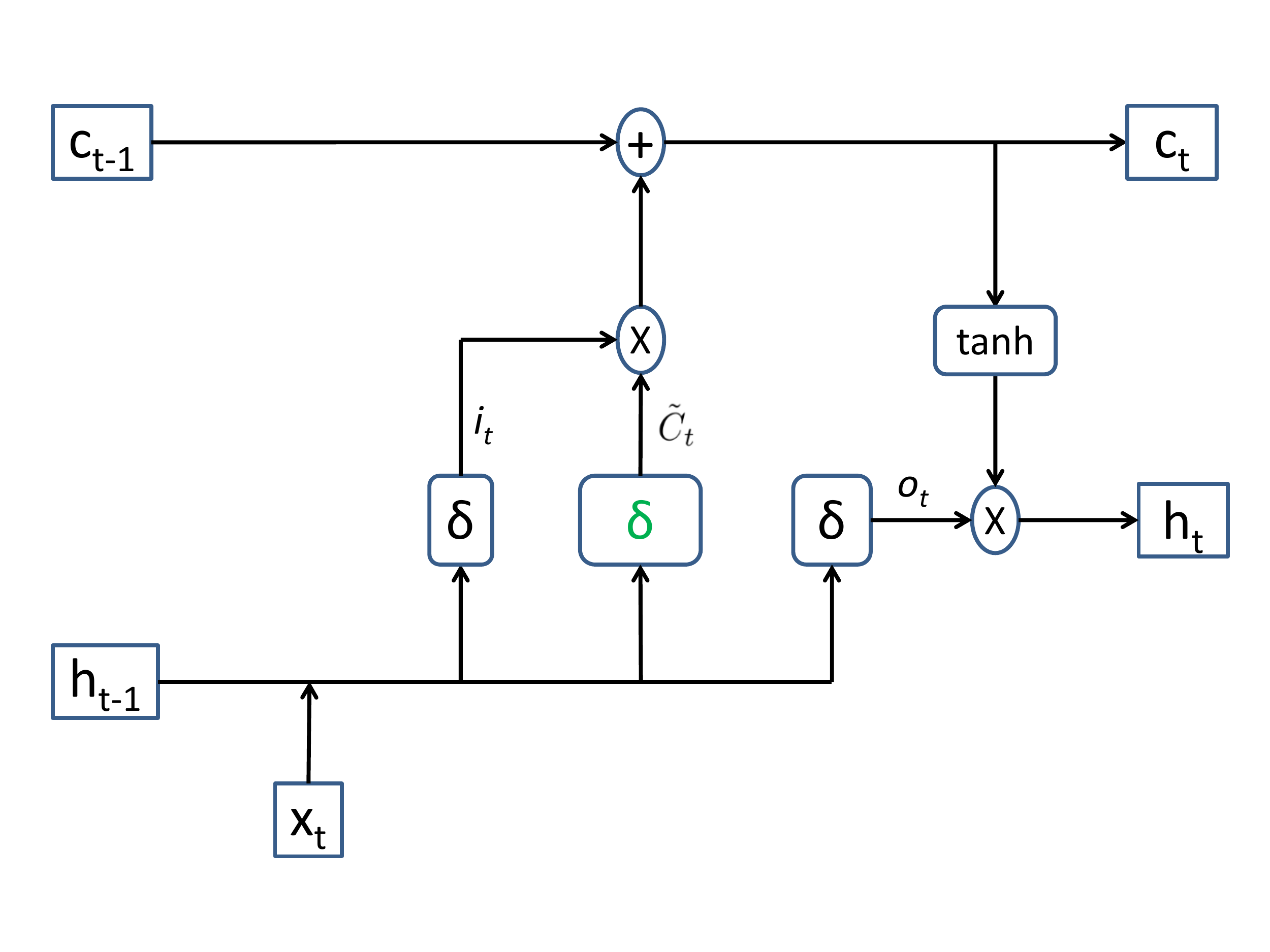}
	\caption{Structure of our proposed LSTM cell without forget gate and the substituted \textit{tanh(.)} activation.}
\end{figure}
Compared to standard LSTM, the proposed LSTM architecture where the forget gate is "forgotten" gains performance effectiveness and computational efficiency. The former is attributed to the fact that the model is made to learn long-term sequential data without suppressing previous information. On the other hand, computational efficiency gains are manifested as the proposed LSTM works with lesser number of parametres (due to elimination of forget gate \textit{f\textsubscript{t}} and its weight \textit{W\textsubscript{f}} and bias \textit{b\textsubscript{f}}). Mathematically, the new LSTM structure is modelled as below:
\vspace*{-0.2cm}
\begin{equation}
	i_t=\delta(W_i\otimes[h \textsubscript{t-1},x_t] + b_i)
\end{equation}
\begin{equation}
	\tilde{C}_t =\delta(W_C\otimes[h \textsubscript{t-1},x_t] + b_C)
\end{equation}
\begin{equation}
	C_t=C\textsubscript{t-1} + \tilde{C}_t
\end{equation}
\begin{equation}
	o_t=\delta(W_o\otimes[h \textsubscript{t-1},x_t] + b_o)
\end{equation}
\begin{equation}
	h_t=o_t\otimes tanh(C_t)
\end{equation}
\subsection{Regularity Metrics}
Regularity score is a common metrics that calculates the uniformity of a test data's behaviour by comparing with a trained model's pattern \cite{chong2017abnormal,hasan2016learning}. Following model training, we carry out performance evaluation by analyzing how well a model can detect abnormal events (unobserved during training) by providing with test data containing anomalous scenes. A well trained and learnt model detects and discriminates events diverging from the learnt pattern. Such models are also capable of easily suppressing false positives that can affect their anomaly detection performance. We calculate a particular test video's regularity score using the reconstruction error \textit{rec\textsubscript{err}}. We compute the reconstruction error using the Euclidean distance (L2-norm) measured from an input test cuboid and its reconstructed sequence (Eq.(\ref{eq:rec_err})).
\begin{equation}
\label{eq:rec_err}
rec\textsubscript{err}(t)=\sqrt{(x(t) - W_m(x(t)))^2}
\end{equation}
where \textit{x(t)} is the \textit{t\textsuperscript{th}} target test frame, \textit{W\textsubscript{m}} stores trained weights of the model, and \textit{W\textsubscript{m}(x(t))} is the \textit{t\textsuperscript{th}} output frame. The reconstruction error of an input video is the average reconstruction error of every input cuboid fed to the model once with batch size N (N is 8 in our setup). The regularity score of a \textit{t\textsuperscript{th}} test frame \textit{reg\textsubscript{score}(t)} is driven using the equation:
\begin{equation}
\label{eq:reg_score}
reg\textsubscript{score}(t) = 1 - {\frac{rec\textsubscript{err}(t) - min(rec\textsubscript{err}(t))}{max({rec\textsubscript{err}(t)})}}
\end{equation}

The expression at the right side calculates irregularity score of test video using volume reconstruction cost of Eq.(\ref{eq:rec_err}). It reduces the reconstruction cost by subtracting the minimum cost of an anomalous frame from each frame's cost. The obtained result is then divided by the most anomalous frame's reconstruction error.
\subsection{Learning Normality}
During training, we learn appearance and motion features of an anomaly-free environment to improve abnormality detection and generalization performance. The core component of our learning process is a trained network that takes dense optical flow data as input and outputs a normality-aware model. This model stores regular appearance and motion information of every pixel in the video frames of the train set. We train the network on a fully unlabelled dense optical flow data. At a single time step, dense optical flow fields of four frame sequences from a video are fed to the network. During model evaluation, the normality-aware patterns learnt from these dense optical flows are capable of effectively detecting and identifying abnormalities deviating from the normalcy.

\subsection{Detecting Anomalies}
We deploy our trained model on sequence of test frames to evaluate its abnormality detection performance. Identifying and detecting an anomalous frame is accomplished by calculating model reconstruction error. Reconstruction error is a popular measurement scheme to determine the presence of anomalous scenes in test frames. In this work, we use frame-level criterion to evaluate performance of a model on test data. We compute True Positive Rate (TPR) and False Positive Rate (FPR) of the observations in test set using Eqs.(\ref{eq:tpr}) and (\ref{eq:fpr}) respectively which are then used to calculate Area Under Curve (AUC) and Equal Error Rate (EER) of the model. \par
\begin{equation}
\label{eq:tpr}
TPR = \frac{TP}{TP + FN}
\end{equation}
\begin{equation}
\label{eq:fpr}
FPR = \frac{FP}{FP + TN}
\end{equation}
where \textit{TP} is the number of true positives, \textit{FN} is the number of false negatives, \textit{FP} is the number of false positives and \textit{TN} is the number of true negatives observed during model evaluation.

\subsection{Generalizing Abnormalities}
The training process yields models that store normal spatio-temporal features of a particular environment (dataset). Abnormality generalization measures the effectiveness and robustness of such context-specific models by evaluating on an entirely different dataset whose environment is not seen during training. We use a similar AUC by EER metrics for performance evaluation of these models on test set. Abnormality generalization results show that our bi-gated LSTM architecture and self-contained network can robustly handle the temporal dynamics of sequential data, extract spatial features, model motion information in long sequences and discriminate a wider range of abnormalities from several datasets. \par

\section{Experimental Setup}
\subsection{Datasets and Modelling}
\subsubsection{Datasets}
We train and evaluate our model on three popular abnormal event detection datasets; CUHK Avenue \cite{lu2013abnormal}, UCSD Ped1 and UCSD Ped2 \cite{li2014anomaly}. These datasets contain videos of distinct scenarios captured using a fixed camera in an outdoor environment. Videos in the training sets of these datasets consist of only normal events, while videos in the test sets are made of both set of normal and abnormal events. Events that are deemed as abnormal in a particular dataset may not be classified in the same category for another dataset; i.e., the abnormality of an event and hence of a pixel or frame depends on the particular environment the dataset is prepared.\\
{\bf CUHK Avenue} is collected at the Avenue of Chinese University of Hong Kong. It is populated with 16 training and 21 test videos split into 15328 frames for training and 15324 for testing each with a resolution of 640 x 360 pixels \cite{lu2013abnormal}. Avenue's test set contains frame-level masks for ground truth annotations. Fourteen distinct events are classified as abnormal in this dataset like loitering, running, romping, pushing a bike, moving towards camera, and throwing a paper or bag.\\
{\bf UCSD} is one of the most challenging datasets for video abnormality detection problem. It consists of video recordings captured from two distinct pedestrian walkways. Videos from the first pedestrian walkway make up Ped1 dataset comprising 34 training and 36 test videos, which are split into frames of 238x158 pixel resolution. On the other hand, Ped2 contains 16 training and 12 test videos whose frames have a resolution of 360x240 pixels \cite{li2014anomaly}. Every video in Ped1 test set is fragmented into 200 frames, whereas each of Ped2’s test set videos are split into a varying number of 120, 150 and 180 frames. Both Ped1 and Ped2 datasets contain frame-level annotation for abnormalities in the test sets. Abnormal events in this dataset include the presence of car, bicycle, wheelchair, skateboard, and walking on grass or moving in wrong direction across a walkway.\par
\subsubsection{Model Implementation}
\subsubsection{Preprocessing}
We generate sparse and dense optical flows for every raw video in the training set of Avenue, Ped1 and Ped2 datasets. We then change these video-form optical flows into frames similar to the setup in the original dataset \cite{li2014anomaly,lu2013abnormal}. These optical flow vectors of the training set and raw videos from the original test set are converted into frames of size 224x224 pixels. Pixels of the optical flow input frames are then scaled down to a range between 0 and 1 so that the frames are on a same scale. The frames containing optical flow information and image content are then changed to gray-scale and normalized to assume a mean value of zero and a variance of one. The input to the network is a cube built by stacking a sequence of optical flow frames with dynamic number of skipping strides. The input cube has a size of $T \times\,224 \times 224 \times 1$ (where \textit{T} is assigned to 4 in all of our experiments). \par 
\vspace{-0.4cm}
\subsubsection{Model Learning}
We separately train our deep network on sparse and dense optical flow data that we prepare for CUHK Avenue and UCSD datasets for sixty epochs. We also train the network on the original Avenue and UCSD datasets for performance comparison (discussed in Section 4.3.). We divide the train sets into two sub sets: 1) a set containing eighty five percent of the training data that is used for training the network; and 2) a set that is made of the remaining fifteen percent which is used for validating the model. We use Adam gradient-based optimizer proposed by Kingma and Ba \cite{kingma2014adam} with a learning rate of 10\textsuperscript{-5} for optimizing the network, and a single batch of size 8. Adam is capable of automatically adjusting the learning rate by reviewing previous model weights, and is computationally cheap and efficient. After training for the required number of epochs, we evaluate each model generated after every epoch with test data. We then choose the best AUC by EER evaluation result produced by the most effective model.
\subsubsection{Model Evaluation}
We evaluate the performance of our developed model using the Area Under Curve (AUC) by Equal Error Rate (EER) metrics, a popular scheme to evaluate the effectiveness of models for such reconstruction based tasks \cite{hasan2016learning,zhao2017spatio}. The Area Under Receiver Operating Characteristics (ROC) Curve (AUC) measures how well a model is capable of discriminating features amongst multiple classes. Higher AUC values indicate better discrimination performance and robustness of models to various classes and unseen environment. The ROC curve is a result of plotting the True Positive Rate (TPR) along False Positive Rate (FPR) across different threshold values. The TPR is plotted on the y-axis and FPR along the x-axis for dynamically varying thresholds. Using these variables, the AUC can be calculated as the integral of TPR by FPR (Eq.(\ref{eq:auc})). The Equal Error Rate (EER) measures the error margin where False Acceptance Rate (FAR) and False Rejection Rate (FRR) reach closer values. The false acceptance rate measures the number of wrong samples accepted as correct, while false rejection rate measures the number of correct samples which are incorrectly rejected. The EER in turn is the FPR value at the point where FPR equals 1 – TPR \cite{sun2017abnormal}.\\
\vspace*{-0.4cm}
\begin{equation}
\label{eq:auc}
AUC = \int_{0}^{1} TPR(z)dz
\end{equation}
where \textit{z} represents the false positive rate.
\subsection{Comparison with State-of-the-Art Methods}
We compare the performance of our method with related works and standard LSTM on video anomaly detection datasets (shown in Table 1). The Standard LSTM method depicts a network built from conventional LSTM cells (whose forget gate is not removed). Performance evaluation shows the results we obtain with a model trained on dense optical flow data. Figure 6 shows AUC-by-EER ROC curves produced by our network (trained on dense optical flows) when evaluated with the original test sets of Avenue, Ped1 and Ped2. Our proposed LSTM cell and the self-contained network yields enhanced detection performance than standard LSTM based models achieving 11.4\% and 12.8\% improvement on AUC and EER measures respectively with Ped2 test set, almost 4\% improvement on both AUC and EER with Ped1 test set. It also gains 2.5\% AUC betterment than standard LSTM based network on Avenue test set. The proposed model shows performance gains when learning on dense optical flow data than learning on videos from the original datasets (discussed in detail in Section 4.3). Our model that is trained on dense optical flow data achieves 0.9\% AUC and 2.0\% EER improvement than a model trained on the original dataset for Ped2 test set. Our dense optical flow based model also detects abnormalities better than sparse optical flow based model achieving 0.4\% AUC and 1.0\% EER gains for Ped2 test set. Similarly, we notice significant performance improvements on Avenue and Ped1 test sets too. The proposed approach also achieves competent and closer detection performance when compared to other related works on UCSD Ped2 test set. Despite the presence of challenging scenarios in these datasets (like camera shakes, illumination variations and low resolution frames), learning our deep LSTM network on dense optical flows results in a robust model capable of effectively discriminating different forms of irregularities.\par
\begin{table}[!h]
\caption{\label{tab1}Performance comparison of the proposed method with related methods and Standard LSTM based network on CUHK Avenue and UCSD datasets.}
\centering
\hskip-0.2cm
\begin{tabular}{|p{3.0cm}|p{.9cm}p{.9cm}|p{.9cm}p{.9cm}|p{.9cm}p{.7cm}|}
\cline{1-2}
\hline
\rule{0pt}{12pt}\multirow{2}{*}{Method} & \multicolumn{2}{c|}{Avenue}                   & \multicolumn{2}{c|}{Ped1}                     & \multicolumn{2}{c|}{Ped2}                     \\  & 
\rule{0pt}{15pt}AUC   & \multicolumn{1}{l|}{EER}        & AUC                   & \multicolumn{1}{l|}{EER}          & AUC                   & EER                   \\ \cline{2-7}\hline \hline
\rule{0pt}{10pt}Chong et al. \cite{chong2017abnormal}& 80.3 & 20.7 & 89.9 & 12.5 & 87.4 & 12.5   \\
Hasan et al. \cite{hasan2016learning}& 70.2 & 25.1 & 81.0 & 27.9 & 90.0 &21.7    \\
Hinami et al. \cite{hinami2017joint}& -- & -- & 69.9 & 35.9 & 90.8 & 17.1   \\ 
Luo et al. \cite{luo2017remembering}& 77.0 & -- & 75.5 & -- & 88.1 & --   \\
Wang et al. \cite{wang2020abnormal}& -- & -- & 65.2 & 21.0  & -- & --     \\
Zhao et al. \cite{zhao2017spatio}& 80.9 & 24.4 & 87.1 & 18.3  & 88.6  &20.9     \\ \hline

\rule{0pt}{10pt}Standard LSTM & 65.2 & 36.3  & 65.2 & 38.1   & 75.6  & 31.4              \\ 

\textbf{Ours (bi-gated LSTM)} & \textbf{67.6} & \textbf{36.2}  & \textbf{69.7} & \textbf{32.2}  & \textbf{87.0} & \textbf{18.7}              \\ \hline

\end{tabular}
\end{table}

\begin{figure}[!h]
	\centering
	\hspace*{-0.2cm}
	\includegraphics[scale=1.0, width=140mm, height=75mm]{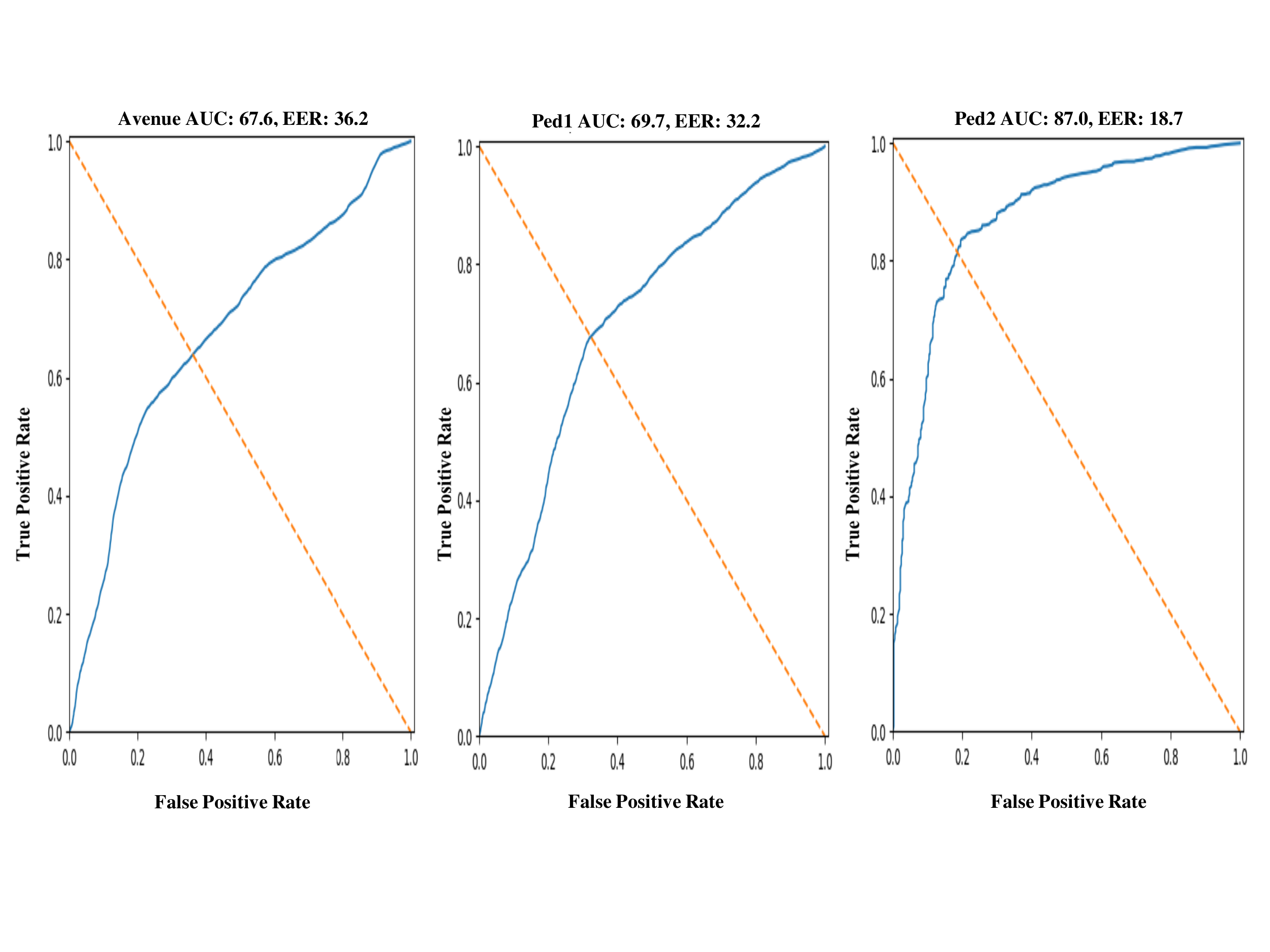}
	\begin{center}
		\vspace*{-1.4cm}
		\caption{ROC curves generated by our network trained on dense optical flows of Avenue, Ped1 and Ped2 dataset respectively (appears better in zoom).}
	\end{center}
\end{figure}

\begin{figure*}[!ht]
	\centering
	\includegraphics[scale=1.5, width=\textwidth, height=110mm]{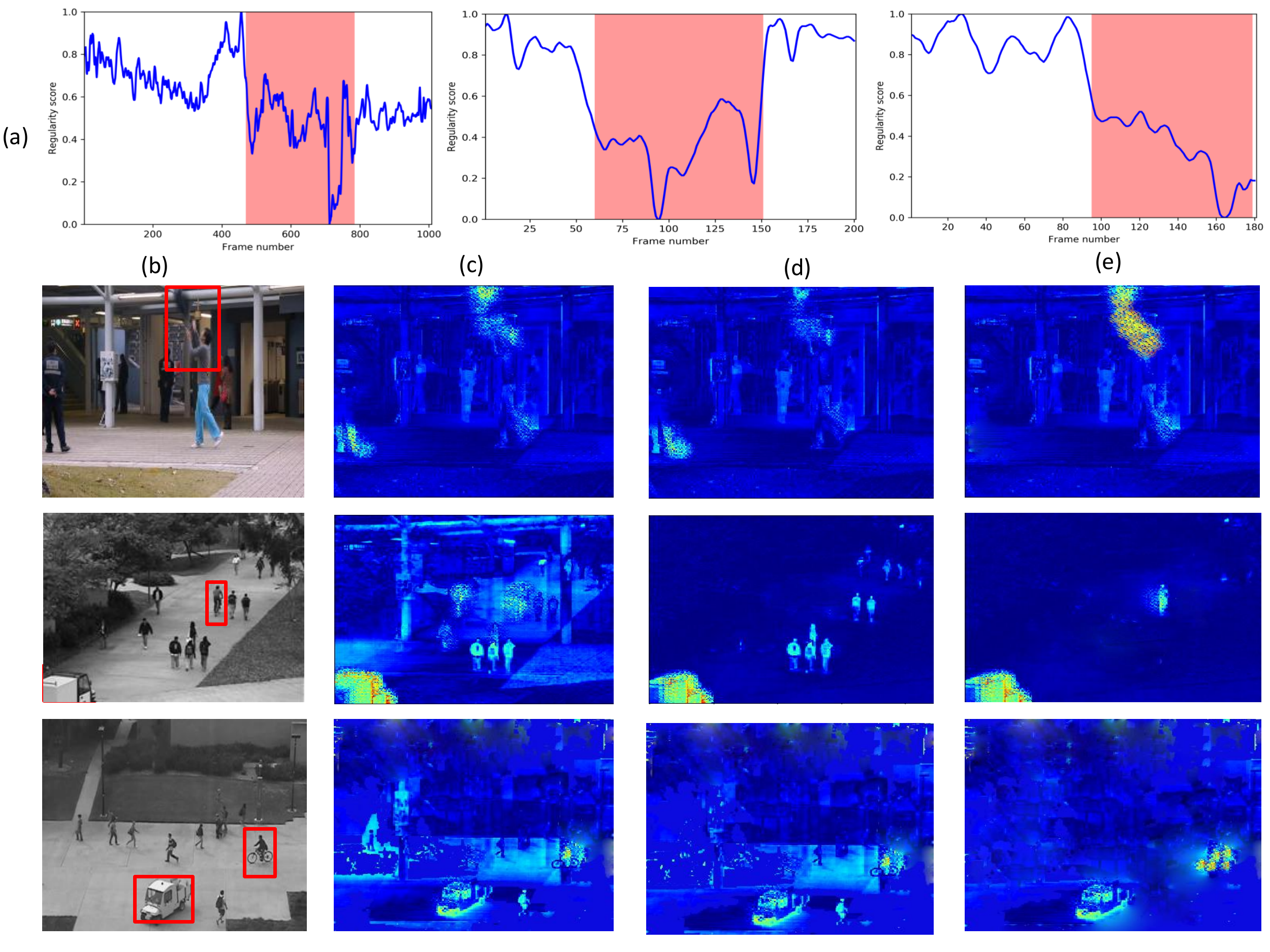}
	\begin{center}
		\vskip-0.6cm
		\caption{\textbf{(a)}(Left-to-right) Regularity score of sample videos from Avenue, Ped1 and Ped2 test sets respectively. \textbf{(b)} (Top-to-bottom) groundtruth annotations from Avenue (throwing), Ped1 (car and biking) and Ped2 (car and biking) test sets respectively. \textbf{(c)} (Top-to-bottom) corresponding qualitative results using standard LSTM. \textbf{(d)} (Top-to-bottom) corresponding qualitative results using LSTM (without input gate). \textbf{(e)} (Top-to-bottom) corresponding qualitative results using our LSTM (without forget gate). Regions highlighted in yellow and red-yellow indicate pixels containing anomalous scenes (appears best in colour).}
	\end{center}
\end{figure*}

We also qualitatively evaluate the performance of our trained model on test sets of AED benchmarks. The qualitative evaluation emphasizes the quantitative results (reported in Table 1) and associates the presence of anomalous scenes with frame behaviour. Figure 7(a) shows regularity score of sample test videos from CUHK Avenue, Ped1 and Ped2 datasets respectively. Figure 7(b-e) show groundtruth abnormalities and the evaluation heatmaps generated using different variants of LSTM based methods. The regions highlighted in dark-yellow and red-yellow show the presence of anomalous events (viz., \textit{throwing a bag} in Avenue, and \textit{car and biking} in both Ped1 and Ped2 datasets) on the specific pixels in the frames.\par
We can witness from the figures that the proposed method can effectively single-out appearance and motion abnormalities in different environments. Whereas the standard LSTM (Figure 7(c)) and LSTM without input gate (Figure 7(d)) are deceived by various normal appearance and motion information (such as \textit{person standing}, background colour and \textit{person walking}) and treat them as abnormal, our bi-gated LSTM shows robust performance to such challenging scenarios.

\subsection{Ablation Study}
We carry out two forms of ablation study to assess performance of the proposed approach. The first one is input-based ablation where we train our bi-gated LSTM network on different forms of input training data. The second one is component-based ablation where we keep or remove the forget or input gate and train the resulting network on dense optical flow data.\par
\subsubsection{Input-based Ablation}
We perform various set of experiments by supplying different inputs to the proposed network architecture. In addition to providing dense optical flow data as input to the network, we train it on train sets from the original datasets and sparse optical flow data. We then evaluate the performance of these three models with the same test set from the original dataset. Table 2 highlights detection performance of models trained on different train sets. The table tells us that the bi-gated LSTM network performs significantly better when trained on dense optical flows than sparse optical flows and the original dataset.\par
\begin{table}[!h]
	\caption{\label{tab2}Performance comparison by learning the proposed bi-gated LSTM network on different train sets.}
	\centering
	\begin{tabular}{|p{3.5cm}|p{0.8cm}p{0.8cm}|p{0.8cm}p{0.8cm}|p{0.8cm}p{0.8cm}|}
		\hline
		\multirow{2}{*}{Method}             & \multicolumn{2}{c|}{\rule{0pt}{10pt}Avenue} & \multicolumn{2}{c|}{Ped1} & \multicolumn{2}{c|}{Ped2} \\ 
		& \rule{0pt}{12pt}AUC          & EER          & AUC         & EER         & AUC         & EER         \\ \hline
		\rule{0pt}{10pt}Bi-gated LSTM (on original train set)        & 64.5         & 37.2         & 66.9  & 37.0       & 86.1        & 20.7        \\ \hline
		\rule{0pt}{10pt}Bi-gated LSTM (on sparse opt-flow train set) & 62.5         & 37.5         & 68.4        & 34.8        & 86.6        & 19.8        \\ \hline
		\rule{0pt}{10pt}\textbf{Bi-gated LSTM (on dense opt-flow train set)}  & \textbf{67.6}         & \textbf{36.2}         & \textbf{69.7}        & \textbf{32.2}        & \textbf{87.0}        & \textbf{18.7}        \\ \hline
	\end{tabular}
\end{table}
\subsubsection{Component-based Ablation}
We also conduct another set of ablation evaluation on networks built by removing the forget gate or input gate (one at a time) from standard LSTMs. We train these networks on the preprocessed dense optical flow and test them with videos from the original dataset. Table 3 shows performance comparisons of these models and standard LSTM based model. As we can learn from the table, our bi-gated LSTM method outperforms standard LSTM and LSTM without input gate.\par
\begin{table}[!h]
	\caption{\label{tab3}Performance comparison by removing and keeping forget gate and/or input gate from standard LSTMs.}
	\centering
	\begin{tabular}{|p{3.4cm}|p{0.8cm}p{0.8cm}|p{0.8cm}p{0.8cm}|p{0.8cm}p{0.8cm}|}
		\hline
		\multirow{2}{*}{Method}             & \multicolumn{2}{c|}{\rule{0pt}{10pt}Avenue} & \multicolumn{2}{c|}{Ped1} & \multicolumn{2}{c|}{Ped2} \\ 
		& \rule{0pt}{12pt}AUC          & EER          & AUC         & EER         & AUC         & EER         \\ \hline
		\rule{0pt}{10pt}Standard LSTM  & 65.2 & 36.3  & 65.2 & 38.1  & 75.6  & 31.4   \\ \hline
		\rule{0pt}{10pt}LSTM (w/o input gate) & 65.8 & 35.9 & 66.1 & 37.0 & 85.5 & 22.0 \\ \hline
		\rule{0pt}{10pt}\textbf{Bi-gated LSTM} & \textbf{67.6} & \textbf{36.2} & \textbf{69.7} & \textbf{32.2} & \textbf{87.0} & \textbf{18.7}        \\ \hline
	\end{tabular}
\end{table}
\subsection{Generalization Study}
In this section, we discuss the experiments we conduct to assess the generalization capability of our trained models to different datasets. The specific environment on which CUHK Avenue and UCSD datasets are prepared is entirely unrelated with one another. This makes model generalization across datasets a challenging endeavour as both the training environment and the abnormal events in the respective test sets are mostly unique to their datasets. We evaluate if a model trained on a particular dataset is able to effectively identify and detect anomalies when tested with test data from another dataset. More specifically, a model learnt from the dense optical flow data of a particular dataset is evaluated with video frame data from another dataset. For each dataset, we pick the model that yields the best performance result (reported in Table 1) and evaluate its generalization performance on a dataset from different category. We carry out performance comparison with standard LSTM, LSTM without input gate and our bi-gated LSTM architecture for generalization task. Empirical results show that the proposed LSTM based network discriminates abnormalities from another datasets better than standard LSTM and LSTM (without input gate).\par
Tables 4, 5 and 6 show performance result of models trained on a particular dataset and tested with videos from another dataset. Table 4 evaluates models trained on Avenue train set with test sets from Ped1 and Ped2 datasets. Table 5 evaluates models trained on Ped1 train set with test sets from Avenue and Ped2. Table 6 evaluates models trained on Ped2 train set with test sets from Avenue and Ped1 test sets. These results show that the rich appearance and motion contents that our approach can learn enables it to acclimate well to different environments. This in turn helps the trained model to perform consistently in detecting and identifying context-specific anomalies better as enough details are acquired during training.\\
\vspace{-0.7cm}
\begin{table}[!h]
	\centering
	\caption{\label{tab4} Generalization performance of Avenue trained model on Ped1 and Ped2 test sets.}
	\begin{tabular}{|p{3.65cm}|p{.7cm}p{.7cm}|p{.7cm}p{.7cm}|}
		\hline	\rule{0pt}{12pt}\multirow{2}{*}{Method} &\multicolumn{2}{c|}{Ped1} 
		&\multicolumn{2}{c|}{Ped2}    \\  & \rule{0pt}{12pt}AUC   & \multicolumn{1}{l|}{EER}        & AUC                   & \multicolumn{1}{l|}{EER}                    \\ \cline{2-5}\hline \hline
		\rule{0pt}{15pt}Standard LSTM & 62.6 & 38.9 & 84.1 & 21.8   \\ 
		LSTM (w/o input gate) & 65.2 & 36.7 & 85.0 & 21.8 \\
		\textbf{Bi-gated LSTM} & \textbf{72.9} & \textbf{31.4} & \textbf{85.8} & \textbf{21.4}   \\ \hline
	\end{tabular}
\end{table}
\vspace{-1cm}
\begin{table}[!h]
	\centering
	\caption{\label{tab5} Generalization performance of Ped1 trained model on Avenue and Ped2 test sets.}
	\begin{tabular}{|p{3.65cm}|p{.7cm}p{.7cm}|p{.7cm}p{.7cm}|}
		\hline	\rule{0pt}{12pt}\multirow{2}{*}{Method} &\multicolumn{2}{c|}{Avenue} &\multicolumn{2}{c|}{Ped2}    \\  & \rule{0pt}{12pt}AUC   & \multicolumn{1}{l|}{EER}        & AUC                   & \multicolumn{1}{l|}{EER}                    \\ \cline{2-5}\hline \hline
		\rule{0pt}{15pt}Standard LSTM & 64.8 & 36.3 & 85.2 & 22.7   \\ 
		LSTM (w/o input gate) & 63.4 & 41.4 & 84.2 & 21.7 \\
		\textbf{Bi-gated LSTM} & \textbf{68.2} & \textbf{34.8} & \textbf{86.6} & \textbf{19.8}   \\ \hline
	\end{tabular}
\end{table}
\begin{table}[!h]
	\centering
	\caption{\label{tab6} Generalization performance of Ped2 trained model on Avenue and Ped1 test sets.}
	\begin{tabular}{|p{3.65cm}|p{.7cm}p{.7cm}|p{.7cm}p{.7cm}|}
		\hline	\rule{0pt}{12pt}\multirow{2}{*}{Method} & 		\multicolumn{2}{c|}{Avenue} &\multicolumn{2}{c|}{Ped1}    \\  &  \rule{0pt}{12pt}AUC   & \multicolumn{1}{l|}{EER}        & AUC                   & \multicolumn{1}{l|}{EER}                    \\ \cline{2-5}\hline \hline
		\rule{0pt}{15pt}Standard LSTM & 52.9 & 46.6 & 53.1 & 46.6   \\ 
		LSTM (w/o input gate) & 54.3 & 46.2 & 58.8 & 42.4 \\
		\textbf{Bi-gated LSTM} & \textbf{60.3} & \textbf{43.0} & \textbf{66.8} & \textbf{35.8}   \\ \hline
	\end{tabular}
\end{table}
\subsection{Assessing Computational Efficiency}
Our bi-gated LSTM based network not only achieves better performance effectiveness on abnormal event detection datasets, it is also computationally inexpensive taking less training and testing time with fewer memory consumption than other LSTM based networks. We experiment all training and testing processes on a single NVIDIA GeForce GTX 1080 GPU with Keras \cite{chollet2015keras} on Tensorflow \cite{tensorflow2015-whitepaper} backend.
\subsubsection{Training Efficiency}
We measure the training efficiency of our network by assessing the per-epoch training time taken to learn a model on all the three datasets. Tables 7 and 8 summarize training efficiency of our proposed network and compare it with related LSTM based networks. As the time taken to generate a model snapshot varies for every epoch, we use the maximum and minimum per-epoch training time to evaluate the training efficiency of our network. We can notice from the tables that the proposed method is efficient for training task than standard and LSTM without input gate networks (Table 7). In addition, learning our network on dense optical flows takes lesser time than training on video frames from the original train sets (Table 8).
\vspace{-0.6cm}
\begin{table}[h]
	\centering
	\caption{\label{tab7}Maximum/minimum per-epoch training time of LSTM variants on AED benchmarks. All methods trained on the preprocessed dense optical flows.}
	\begin{tabular}{|p{3.3cm}|p{1.8cm}|p{1.8cm}|p{1.8cm}|}
		\cline{1-2}
		\hline
		\rule{0pt}{12pt}{Method} & {Avenue} & {Ped1}  & {Ped2}     \\ \cline{2-4}\hline
		\rule{0pt}{12pt}Standard LSTM& 3944/3823 sec & 2667/2364 sec  & 1331/795 sec  \\ \hline
		\rule{0pt}{12pt}LSTM (w/o input gate)& 3514/3210 sec & 1758/1448 sec &  741/536 sec  \\ \hline
		\rule{0pt}{12pt}\textbf{LSTM (w/o forget gate)} & \textbf{3412/3007 sec} & \textbf{1515/1132 sec} & \textbf{564/518 sec}	\\ \hline
	\end{tabular}
\end{table}
\begin{table}[!h]
	\centering
	\caption{\label{tab8}Bi-gated LSTM's maximum/minimum per-epoch training time on different training data. \textbf{\textsuperscript{(*)}} shows our model trained on the original train set, \textbf{\textsuperscript{(**)}} indicates our model trained on dense optical flow data.}
	\begin{tabular}{|p{1.7cm}|p{2.3cm}|p{2.3cm}|p{2.3cm}|}
		\cline{1-2}
		\hline
		\rule{0pt}{12pt}{Method} & {Avenue} & {Ped1}  & {Ped2}     \\ \cline{2-4}\hline
		\rule{0pt}{12pt}\textbf{Ours\textsuperscript{(*)}}& 3653/3310 sec 	& 2133/1624 sec & 1704/549 sec	\\ \hline
		\rule{0pt}{12pt}\textbf{Ours\textsuperscript{(**)}}&\textbf{3412/3007 sec}   & \textbf{1515/1132 sec}   & \textbf{564/518 sec} \\ \hline
	\end{tabular}
\end{table}
\subsubsection{Testing Efficiency}
We also analyze the test efficiency of our network and the trained model by measuring the number of test frames evaluated in one second (on an environment stipulated in Section 4.5). We compare our model's test efficiency with other LSTM variants and a model trained on the original train sets. Tables 9 and 10 highlight performance efficiency comparisons on CUHK Avenue and UCSD datasets. As presented in the tables, the proposed model attains best test efficiency than other LSTM variants due to the absence of forget gate reducing testing overheads (Table 9). This makes it convenient and easily deployable on portable devices for real-time video anomaly detection. We also show that bi-gated LSTM gains test efficiency improvements when trained on dense optical flows than on the original train sets (Table 10).
\begin{table}[!h]
	\centering
	\caption{\label{tab9} Comparison of test efficiency with variants of LSTM based networks on AED datasets. All methods trained on dense optical flows.}
	\vspace{-0.4cm}
	\begin{tabular}{|p{4.3cm}|p{2cm}|p{2cm}|p{2cm}|}
		\hline
		\multicolumn{1}{|c|}{\multirow{2}{*}{Method}} & \multicolumn{3}{c|}{\rule{0pt}{12pt}Dataset}                                          \\ \cline{2-4} 
		\multicolumn{1}{|c|}{}                        & \multicolumn{1}{c|}{\rule{0pt}{12pt}Avenue}     &\multicolumn{1}{c|} {Ped1}                         & \multicolumn{1}{c|}{Ped2}    \\ \hline
		\rule{0pt}{12pt}Standard LSTM     & \multicolumn{1}{c|}{5.2 fps} &  \multicolumn{1}{c|}{6.3 fps}  & \multicolumn{1}{c|}{5 fps}  \\ \hline
		\rule{0pt}{12pt}LSTM (w/o input gate) & \multicolumn{1}{c|}{6.8  fps} &  \multicolumn{1}{c|}{7.0 fps} &\multicolumn{1}{c|}{6.0 fps}    \\ \hline
		\rule{0pt}{12pt}\textbf{LSTM (w/o forget gate)} & \multicolumn{1}{c|}{\textbf{7.0 fps}}  &  \multicolumn{1}{c|}{\textbf{7.2 fps}} &\multicolumn{1}{c|}{\textbf{6.2 fps}}    \\ \hline
	\end{tabular}
\end{table}
\begin{table}[!h]
	\caption{\label{tab10} Testing efficiency of Bi-gated LSTM model trained on different input. \textbf{\textsuperscript{(*)}} shows our model trained on the original train set, \textbf{\textsuperscript{(**)}} indicates our model trained on dense optical flow data.}
	\centering
	\begin{tabular}{|p{2.8cm}|p{1.5cm}|p{1.5cm}|p{1.5cm}|} \hline
	\multicolumn{1}{|c|}{\multirow{2}{*}{Method}} & \multicolumn{3}{c|}{\rule{0pt}{12pt}Dataset}
	\\ \cline{2-4}
	\multicolumn{1}{|c|}{}   & \multicolumn{1}{c|}{\rule{0pt}{12pt}Avenue}     &\multicolumn{1}{c|}{Ped1}                         & \multicolumn{1}{c|}{Ped2}    \\ \hline \rule{0pt}{12pt}Ours\textsuperscript{(*)}             & \multicolumn{1}{c|}{5 fps}  &  \multicolumn{1}{c|}{5 fps} &\multicolumn{1}{c|}{4.1 fps}    \\ \hline
	\rule{0pt}{12pt}\textbf{Ours\textsuperscript{(**)}} & \multicolumn{1}{c|}{\textbf{8.2 fps}} & \multicolumn{1}{c|}{\textbf{8.2 fps}} & \multicolumn{1}{c|}{\textbf{7.2 fps}} \\ \hline
	\end{tabular}
\end{table}
\section{Conclusion}
In this work, we introduce a bi-gated LSTM structure for abnormal event detection and generalization task whose forget gate is removed. We present a self-contained, end-to-end network based on this LSTM cell. The experiments we conduct on CUHK Avenue and UCSD datasets show the effectiveness and efficiency of the proposed LSTM. We show that the proposed LSTM cell gains performance effectiveness and computational efficiency when trained on dense optical flows significantly improving handling of appearance and motion irregularities. Comparative studies indicate the betterment of our method than standard and other variants of LSTMs. Our method also attains performance competitive to the state-of-art methods on UCSD Ped2 dataset. \par
Future works may consider improving performance limitation of the proposed method for Avenue and Ped1 datasets by coupling with other network structures. In addition, pixel-level evaluation and enhancing generalizability of the trained models can also be researched in future works.

\bibliographystyle{abbrv}
\bibliography{mybib}
\end{document}